# Machine Learning-Based Automated Assessment of Intracorporeal Suturing in Laparoscopic Fundoplication

Shekhar Madhav Khairnar MS[1,] Huu Phong Nguyen PhD[1], Alexis Desir MD[1], Carla Holcomb MD[1], Daniel J. Scott MD[1], Ganesh Sankaranarayanan PhD[1]*

[1] Department of Surgery, University of Texas Southwestern Medical Center, Dallas, Texas, USA

*Corresponding author:

Ganesh Sankaranarayanan PhD

Associate Professor

Department of Surgery

UT Southwestern Medical Center

E-mail:

Ganesh.Sankaranarayanan@utsouthwestern.edu

Presented as an oral presentation at the Association for Surgical Education Annual Meeting, Orlando, FL

Statements and Declarations

- Funding this work was supported by the National Institute of Health/National Institute of Biomedical Imaging and Bioengineering grant #R01EB025247

- Mr. Shekhar Madhav Khairnar, Drs. Huu Phong Nguyen, Alexis Desir, Carla Holcomb, Daniel Scott, and Ganesh Sankaranarayanan have no other conflict of interest or financial ties to disclose that are directly or indirectly related to the work submitted for publication.

**Abstract**

**Introduction**

Automated assessment of surgical skills using artificial intelligence (AI) is valuable for trainees to obtain instantaneous feedback. After bimanual tool motions are captured, the derived kinematic metrics have shown to be reliable predictors of performance in laparoscopic tasks. Implementing automated tool tracking assessment requires time-intensive human annotation. We have developed AI-based tool tracking using the Segment Anything Model (SAM) to eliminate the need for human annotators. Here we describe a study to evaluate the usefulness of our tool tracking model in automated assessment of performance in a laparoscopic suturing task of the fundoplication procedure.

**Method**

An automated tool tracking model was applied to recorded videos of Nissen fundoplication on porcine bowel. The participating surgeons were grouped into novices (PGY1-2) and experts (PGY3-5, fellow and attendings). The beginning and the ending of each of the suturing steps were segmented and the motions of the left and right tools were extracted. A low-pass filter with a cut-off frequency of 24 Hz was then applied to remove any noise. Automated assessment of performance was implemented using both supervised and unsupervised models, and an ablation study was performed to compare the performance. For the supervised learning model, kinematic features included root mean square (RMS) velocity, RMS acceleration, RMS jerk, total path length, and Bimanual Dexterity in pixel coordinates (x, y) that were extracted and analyzed using Logistic Regression, Random Forest Classifier, Support Vector Classifier, and XGBoost. We further analyzed the performance with a reduced set of features selected using a Principal Component Analysis (PCA). For unsupervised learning, a Denoising Autoencoder (DAE) model with classifiers like a 1-D (Convolutional Neural Network) CNN and traditional Machine Learning models were trained for classification.

**Results**

Data was extracted for 28 participants out of an initial cohort of 38, categorized into 9 novices and 19 experts. The approach for supervised learning utilized kinematic features with Principal Component Analysis (PCA) using a Random Forest model (the best model among other Machine Learning models) and obtained an accuracy of $0.795 \pm 0.065$ and an F1 score of $0.778 \pm 0.071$. The approach for unsupervised learning employed a 1-D CNN achieved the best results with an accuracy of $0.817 \pm 0.108$ and an F1 score of 0.806 ± 0.110. This second approach is superior as it eliminates the need to compute kinematic features.

**Conclusion**

We successfully demonstrated an AI model for automated classification of performance independent of human annotation of surgical videos that can be used to enhance surgical skill evaluation.

# 1 INTRODUCTION

Surgeon technical skills have shown to directly affect clinical outcomes [1-3]. Effective surgical skill assessment is crucial to ensure patients' safety and reduce clinical errors [4]. Traditional methods of skill assessment in minimally invasive surgery and increasingly more in open surgical procedures, typically rely on video-based assessment (VBA) by expert surgeons[5]. Though based on longstanding educational frameworks, this approach is prone to variation in judgment, sensitive to subjective interpretation, which can affect the reliability of the assessment [6]. Furthermore, VBA is time-consuming and expensive, requiring numerous human hours for video processing alone [7]. Additionally, video review and grading require the involvement of expert surgeons, which can significantly delay the process. To overcome this, objective surrogate measures of skill assessment such as those based on tool motion[8-11], eye tracking[12], physiological measures such as heartrate variability (HRV)[13], and measuring brain activity[14, 15] have been validated in simulated setting with limited usage in operating rooms. The exception to this is the use of kinematic measures in robotic surgery where the surgical tool motion can be easily recorded for later analysis without the need for additional sensors[16], though access to this data is limited for research.

Recent advances in computer vision, where visual information is processed using artificial intelligence, typically a deep learning model is emerged as a valuable VBA tool for surgical workflow analysis, performance assessment, feedback and coaching [17-22]. A particularly useful application of this technology is the automatic detection of the presence, location, or movement of surgical tools is crucial for developing a fast and objective surgical evaluation system [23, 24]. Though prior studies [20, 25] have effectively integrated deep learning models to classify surgical proficiency using pre-processed and manually annotated video datasets, such as the Johns Hopkins University-Intuitive Surgical Incorporated gesture and skill assessment working set (JIGSAWS), which has both synchronized kinematics and surgical skill video data, they frequently require significant efforts to prepare and may not fully capture the dynamic nature of surgical skill progression. Thus, existing approaches, although automated to an extent, still rely heavily on segmented video snippets or kinematic data from sensors, which can be intrusive and disrupt the natural flow of surgical procedures.

To overcome these challenges in this work we introduce a novel approach to surgical skill assessment by automating the creation and labeling of training datasets using the Segment Anything Model (SAM) combined with a “You only look once,” version 8 (YOLOv8) model pre-trained on the Cholec80 dataset.

This integration enhances the precision of tool tracking in surgical videos and significantly reduces the labor-intensive process of manual data annotation, addressing a major bottleneck in surgical skill analysis. We tested our approach by automatically assessing surgical skill in a simulated laparoscopic fundoplication using only the motion-based features extracted directly from the video.

By automating this process, we provide a scalable, efficient, and less subjective method for assessing surgical skills. Second, we compared methods for assessing laparoscopic suturing skills in supervised and unsupervised learning using the data collected during laparoscopic fundoplication. Through an ablation study, we aimed to provide recommendations on the best approach for evaluating suturing performance. The first approach (used for supervised learning) involved directly analyzing kinematic features including root mean square (RMS) Velocity, RMS Acceleration, RMS Jerk, Path Length, and Bimanual Dexterity. The second approach (used for unsupervised learning) involved converting the original 1-D data (x-axis and y-axis) into 2-D images and employed a Denoising Autoencoder (DAE) to indirectly learn features. Third, for the purposes of this research, we compiled a new and comprehensive fundoplication dataset.

## 2 MATERIALS AND METHODS

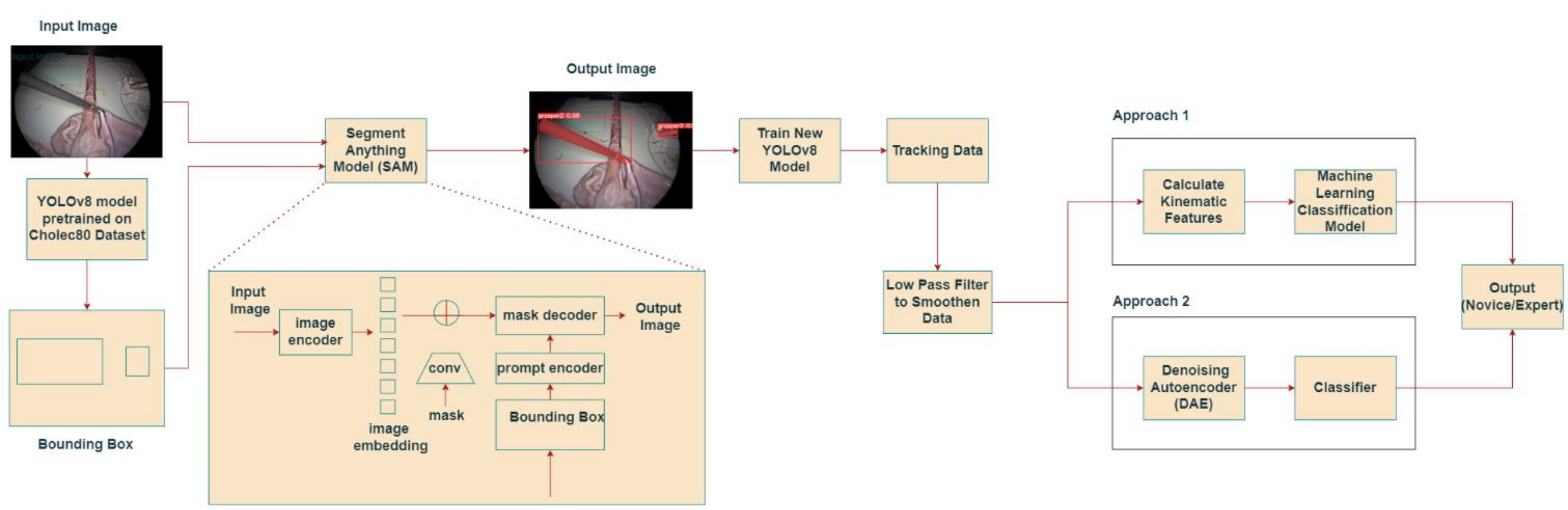

**Fig. 1** Overall architecture of our automated laparoscopic surgical skill assessment with SAM and YOLOv8 integration.

In this section, we begin by discussing the collection of our fundoplication dataset. We then delve into the preprocessing stage, where we have uniquely automated the annotation process by leveraging a combination of YOLOv8 model and SAM. Annotation is essential for training AI models, yet manual annotation is time-consuming. Initially, we considered using either YOLOv8 or SAM for label generation, but each presented a challenge. The YOLOv8 model did not produce bounding boxes with the accuracy needed for subsequent analysis. The SAM required manual prompts, which contradicted our goal of full automation. Therefore, we combined the two approaches: YOLOv8 provides approximate bounding boxes, which SAM then uses for generating precise segmentations. This hybrid method eliminates the need for manual annotation while achieving high accuracy. Next, we explain our methodology for uncovering patterns in the dataset through two distinct approaches. The first approach involves using Machine Learning models to directly analyze the kinematic features data. The second approach converts the original 1-D data into 2-D images and uses a deep learning DAE model to indirectly learn features. An overview of the architecture of our system can be seen in Figure 1. The process begins with an input image taken from a surgical video. A pre-trained AI model called YOLOv8, which was trained using a Cholec80surgery dataset, is used to detect surgical tools by drawing bounding boxes around them. These detected tool areas are passed to another powerful model called Segment Anything Model (SAM), which carefully outlines the shapes of the tools in the image. This helps in precisely identifying the tools in complex surgical scenes.

The combination of YOLOv8 and SAM allows the system to create clean and accurate segmentations of the tools. Using these refined images, a new YOLOv8 model is trained to specialize in detecting and tracking surgical instruments throughout the entire fundoplication procedure. This tool motion data is further processed using a low-pass filter to remove any noise.

The smoothened tool motion data is then used to classify surgical skill into a novice or expert category using two approaches. In approach 1, kinematic features are extracted from the motion data and fed into a machine learning classification model. In approach 2, a Denoising Autoencoder (DAE) model was trained in an end-to-end fashion to automatically extract features using the input kinematic data and use it for classification. Detailed implementation of the tool tracking and classification models are further explained below in this section.

All experiments were conducted on a high-performance workstation equipped with an AMD Ryzen Threadripper PRO 5955WX 16-core processor (4.0 GHz), 128 GB of RAM, and two NVIDIA RTX 4090 GPUs (24 GB VRAM each).

### 2.1 Dataset

The study involved high-definition video recordings of 38 subjects who performed laparoscopic fundoplication up to two times on a porcine model. This was part of the NIH funded project to develop a simulator for training in laparoscopic fundoplication, where task-specific assessment metrics were validated using a porcine fundoplication model performed at the simulation center at UT Southwestern Medical Center [26]. The recordings captured the bimanual manipulations of participants categorized as novices (PGY1-2), intermediates (PGY3-5, fellows) and experts (attendings). The demographics, surgical and simulation training experience of the participants are reported in our previous study[26].

Initially, the dataset consisted of 60 videos from 38 subjects. We used a total of 49 videos, chosen for their camera stability and clear view of the surgical field, and discarded 11 videos due to incomplete tasks or excessive camera movements. The reduced dataset was further classified into two groups with novices (PGY 1-2, n = 9) and experts (PGY 3-5, fellow and attending, n = 19).

## 2.2 Data pre-processing

### 2.2.1 Tool segmentation using YOLOv8

We utilized the YOLOv8 model to generate bounding boxes (a rectangular box around the object of interest, that is used to define the location of an object within an image) for surgical tools from the original images. YOLOs were adopted mainly due to their faster inferences than two-stage detectors [27].

This model, initially pre-trained on the common objects in context (COCO) dataset. The Microsoft COCO dataset is a large-scale benchmark for computer vision tasks such as object detection, instance segmentation, key point detection, and image captioning. It contains around 330k images with over 2.5 million annotated object instances spanning 80 common object categories. Each object is marked with a tight bounding box and an instance-level segmentation mask, and the "person" class includes more than 250k detailed key point annotations for human pose estimation. Additionally, COCO provides five human-written captions per image, offering rich descriptions for language-vision research. The model was fine-tuned for surgical instrument detection using the Cholec80 dataset (the dataset contains 80 videos of laparoscopic cholecystectomy surgeries annotated with surgical phase labels for assessing surgical phase recognition and tool detection models.) [28, 29], annotated with tools like graspers, hooks, and scissors. For our specific application focusing on graspers and needle drivers, the model was adjusted to detect these tools with high precision, using annotations converted into COCO format using LabelMe software [30].

### 2.2.2 Automatic annotation via SAM

After the first YOLOv8 phase, SAM [31] was utilized to improve the tool segmentation further. The extracted frames (input images) along with bounding boxes were then passed to SAM to generate masks. In detail, an image is encoded to capture essential features, and bounding box coordinates (input prompts) are encoded to highlight areas of interest. These encodings are processed by a mask decoder, which also receives a preliminary mask from a convolutional layer for refinement. The resulting refined masks are overlaid on the original image (for more information, see [31]).

### 2.2.3 Train the YOLOv8 model

Later, we retrained the YOLOv8 model to generate bounding boxes with better precision. We extracted 200 frames from each of the 49 videos of 28 subjects with equal number of frames sampled from the start, middle, and end segments of each videos for training the YOLOv8 segmentation model. Of these, 33 videos were used for training, with 965 images annotated to highlight surgical tools and maneuvers, and an additional 31 background images for model training. For validation, 130 images were extracted from 5 videos, and for testing, 306 images were included from 11 videos. In total, only 1401 frames that displayed high-quality segmentation with accurately generated masks were selected.

## 2.3 Data filtering

After training the segmentation model, we extracted the tool motion data in 2-D (x-axis and y-axis) and calculated the position of the tooltip. These data were then divided into left and right tools and then further divided into four parts, namely suturing 1 (S1), S2, S3 and S4 for each tool. Thus, each video had at least 1 suture and some participants did 2 sutures while some did 3 or 4. Although S4 was recorded in a few cases, it was excluded from further analysis because it was performed by only 2 or 3 subjects, whereas S1, S2, and S3 were common across all subjects. Finally, we used the tool motion data for further analysis. To remove high frequency noise from the data, we passed the tool motion data through a low-pass Butterworth filter.

## 2.4 Data Analysis

### 2.4.1 Machine Learning Model

For supervised learning, after filtering, we computed several kinematic features that included RMS Velocity, RMS Acceleration, RMS Jerk, Path Length, and Bimanual Dexterity (bimanual dexterity as a measure of coordination between the left and right tools) for each video. Thus, we had 141 samples (from 9 novice and 19 expert subjects) for a Machine Learning model training for classification of expert and novice subjects. There were 8 features (6 RMS values, plus Bimanual Dexterity and Total Path Length that were combined for both tools). We dropped the acceleration feature since it was highly correlated with jerk.

For training Machine Learning models, we did a 10-fold cross-validation stratified split. In our study, we implemented various data transformation techniques, including Power Transformer, Robust Scaler, Minmax Scaler, and Standard Scaler. Subsequently, we conducted two distinct analyses utilizing Machine Learning models. In the first analysis, we applied PCA to select the most important features with n = 3 components, which accounted for 84% of the variance in the original dataset. The second analysis was performed without PCA. For both analyses, we employed several classifiers: Random Forest [32], XGBoost [33], Support Vector Classifier [34], and Logistic Regression.

### 2.4.2 DAE and Classifier Model

In addition to the conventional Machine Learning methods, we explored the use of autoencoder [35] models to capture more important features from the tool trajectory data (x-axis and y-axis).

Autoencoders, particularly effective in unsupervised learning scenarios, are neural networks trained for compressing input data into a condensed format and then reconstructing it back to its original form. This capability is essential for isolating crucial features from raw data without explicit labels. In our research, we employed a DAE (Figure 2), a type of autoencoder designed to enhance feature extraction by first adding noise to the input and then learning to recover the original, uncorrupted data [25,26]. This prevents the autoencoder from just duplicating the input and encourages it to learn more relevant features.

*Autoencoder training*

Our training involved an autoencoder that processed tooltip x-y coordinate data, representing the extreme points of the left and right surgical tools. The encoder received input images of size 224 × 224 × 3 and reduced this data into a lower-dimensional space measuring 128 × 1. We prepared the data by concatenating the suturing segments S1, S2, and S3 for each subject, with the observation that S1 often took longer than the other segments. Consequently, our dataset comprised separate files for the x and y axes, which were further split into left and right, resulting in a total of 196 samples. We processed the tooltip data for the x and y coordinates separately, applying zero padding and creating line plots for these coordinates against the number of frames. The x-axis and y-axis data were then converted into 2-D images.

These 2-D images were used to train DAE model. The DAE model was trained with additional Gaussian noise, with a noise factor of 0.5. This means that random noise from a normal distribution was added to the data, and the resulting values were clipped to the range [0, 1]. After training, we extracted low-dimensional features from the encoder, represented as 128-dimensional arrays. These features were utilized for classification by feeding them directly into Machine Learning models and a 1-D CNN model.

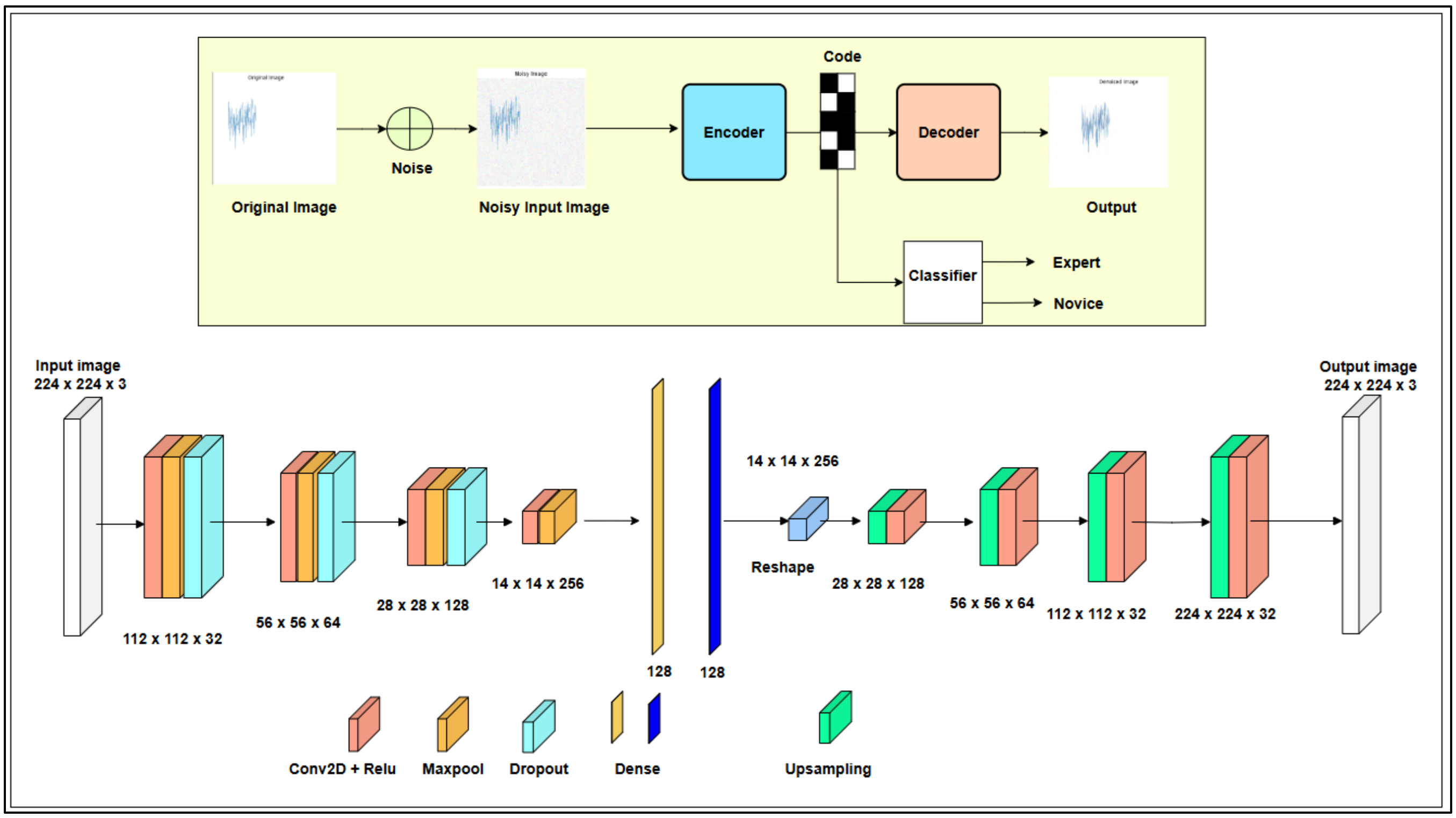

**Fig. 2** Integrated autoencoder workflow for expert and novice classification

## 2.5 Ethical Approval

An institutional review board, the ethics committee of the UT Southwestern Medical Center approved the study design and the use of fundoplication videos (IRB # STU-2021-0151).

# 3 RESULTS

## 3.1 Automated tool tracking

### 3.1.1 Performance of YOLOv8 Cholec80

The YOLOv8 model was pretrained on the Cholec80 dataset, which contains distinct surgical tools/classes (Figure 3). Besides producing bounding boxes, the model also generated masks with the following performance.

The segmentation mask mAP50-95 scores showed varying levels of performance for intersection over union (IoU) thresholds ranging from 50% to 95%. Overall, the model achieved a segmentation mask mAP50-95 score of 0.827 across all classes. For the bipolar surgical tool, the score was 0.828, while the hook tool achieved a higher score of 0.903. The grasper tool had a score of 0.767, and the irrigator tool showed a score of 0.90. The clipper tool had a score of 0.726, and the scissors tool achieved a score of 0.838. Since the detection of graspers is critical, we opted to improve this performance, as described in the next section.

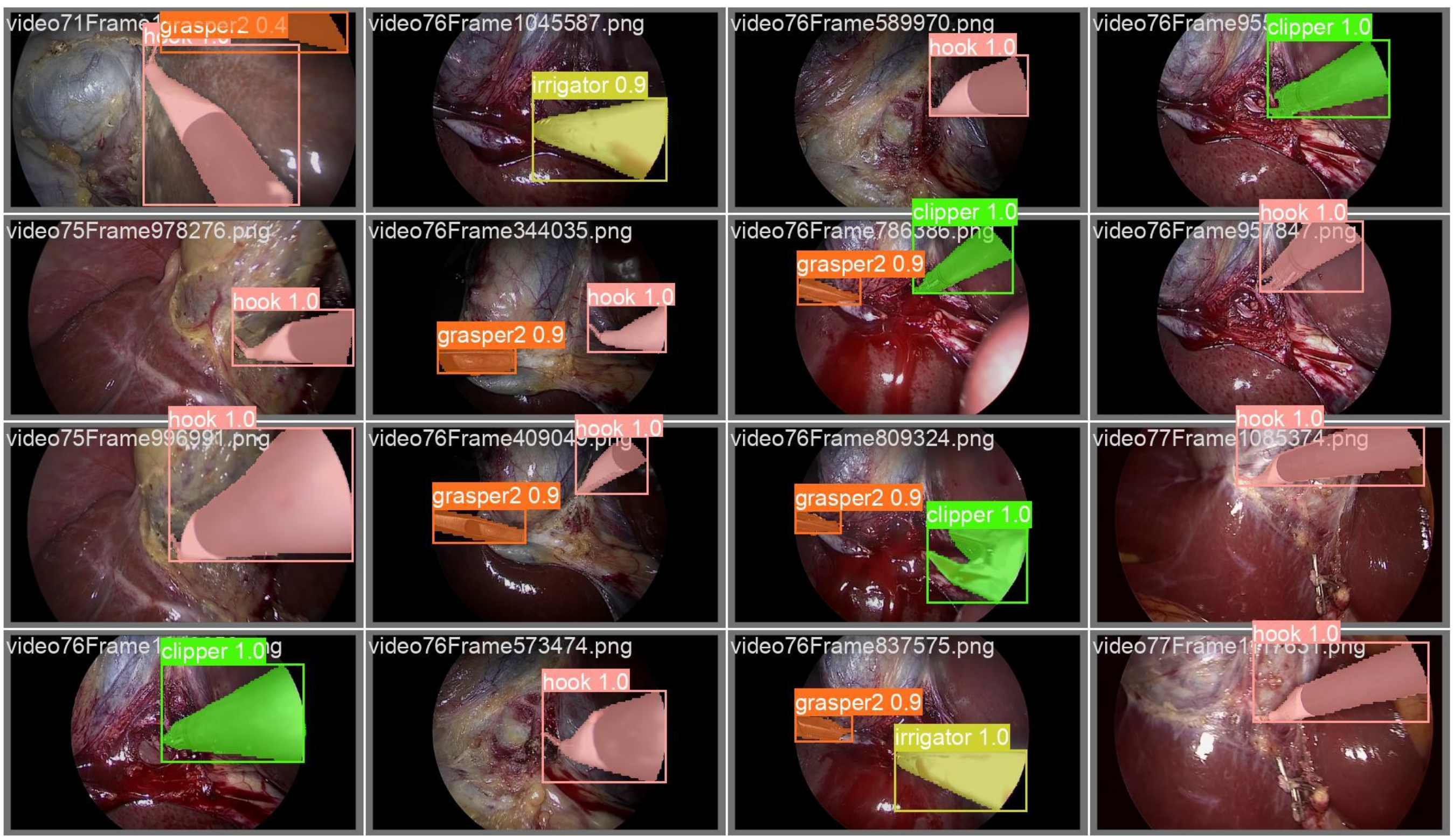

**Fig. 3** YOLOv8 segmentation for various surgical tools on Cholec80 dataset

### 3.1.2 Performance of YOLOv8 on Fundoplication data

From the dataset of 1401 images, the combination of the YOLOv8 and SAM increased the mAP from 0.76 to 0.95 for the mask of the grasper class (Figure 4). This high mAP score for segmentation mask signifies that the model accurately detected and segmented the grasper tool in surgical videos. By leveraging this approach, we eliminated the need for manual annotation, saving both human effort and time.

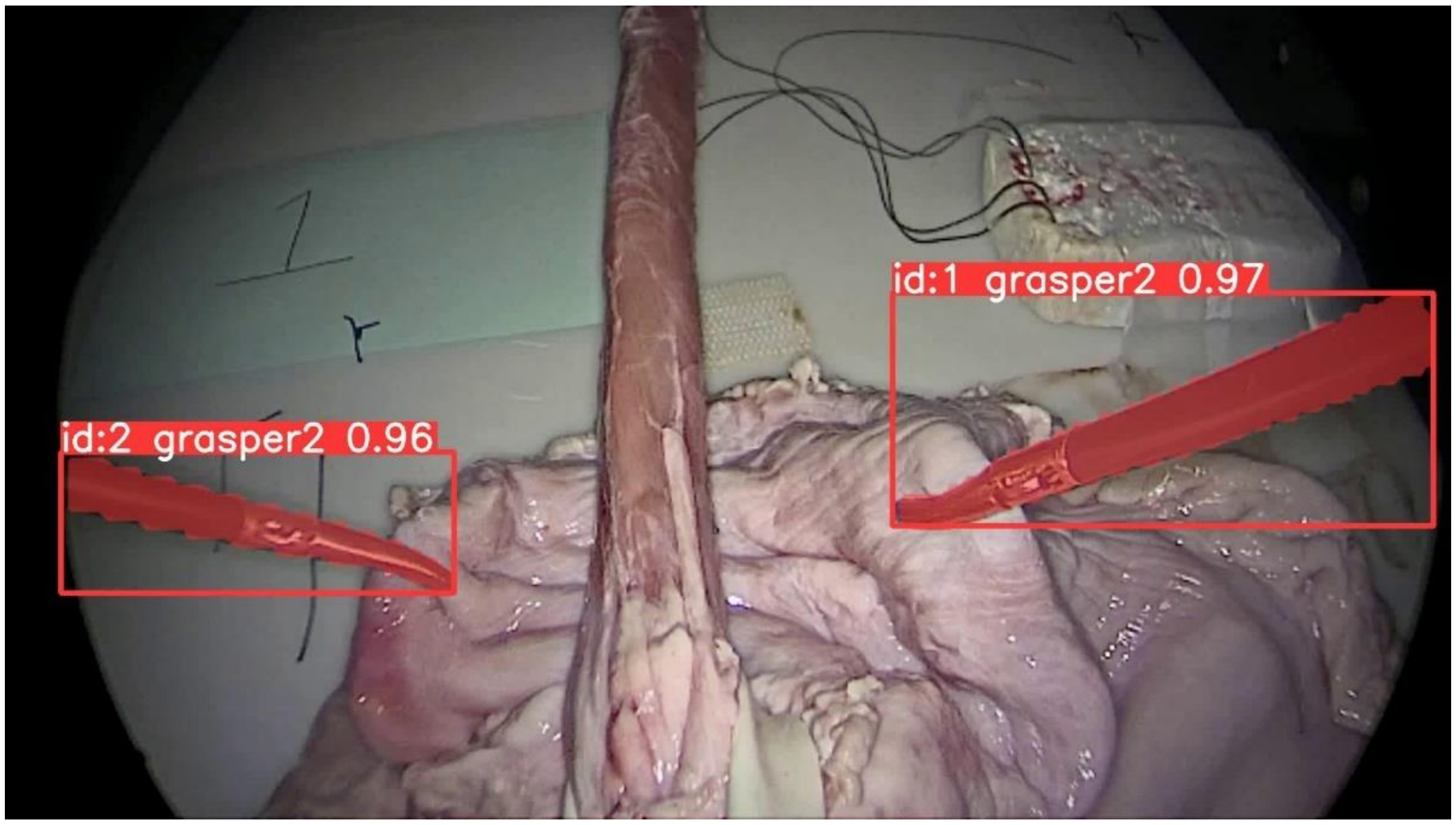

**Fig. 4** YOLOv8 segmentation and tracking for Fundoplication

### 3.1.3 Filtered output

In our approach, we improved the preprocessing of kinematic data by adding an extra layer of processing. Specifically, we applied a filter to smooth the data. As seen in Figure 5, this method effectively reduces noisy data, thereby improving the overall quality and reliability of the data for subsequent analysis.

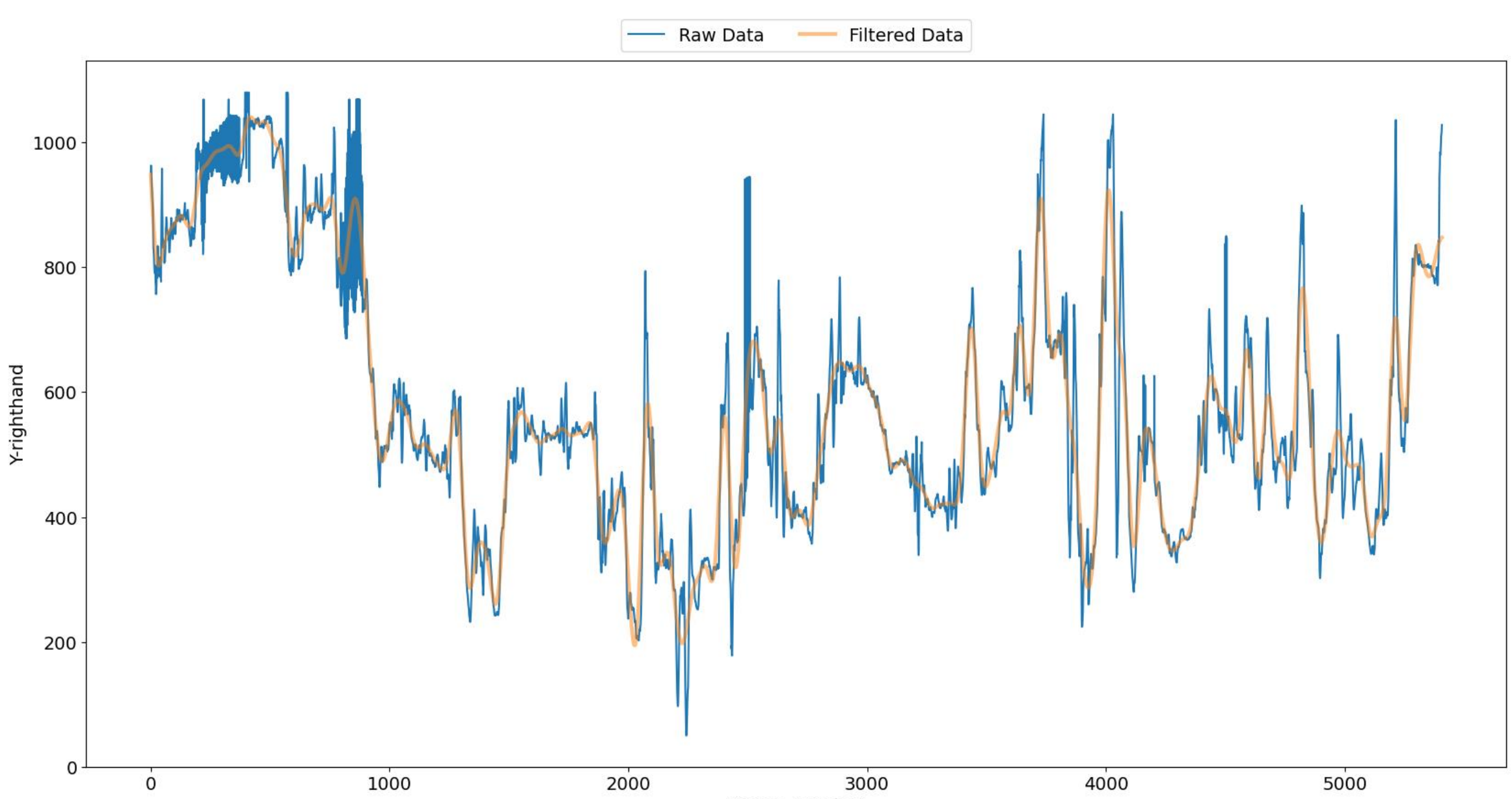

**Fig. 5** Tool Tracking Raw Data versus Filtered Data.

### 3.2 Machine Learning with Kinematic Features

The use of kinematic features is illustrated in the radar plot (Figure 6). Novices took more time than experts to execute movements, had higher jerks, and covered a longer total path length, indicating more erratic and less controlled movements. High bimanual dexterity in experts indicates better coordination between the left and right tools compared to novices.

Since the dataset is unbalanced, i.e., there were 19 experts but only 9 novices, we utilized the weighted loss technique to address the issue. The results show an accuracy of 0.774 and an F1 score of 0.775 without using PCA, and 0.795 and 0.778 using PCA (Table 1). The use of PCA slightly improved performance on most measurement metrics, except for positive predictive value. Please note that only the best model's performance is presented. For a complete set of results for all models, see Table S1 in the Supplementary section.

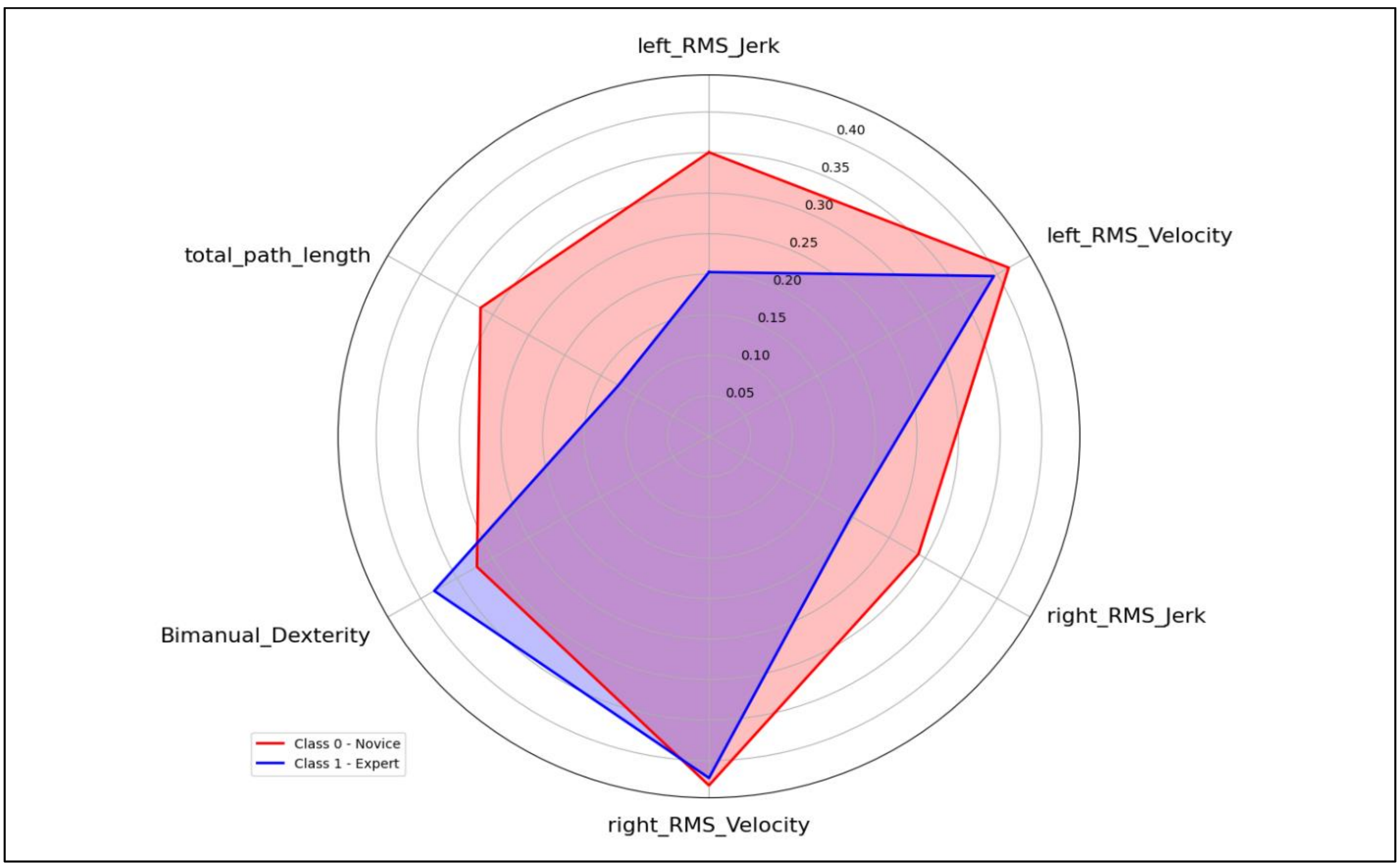

**Fig. 6** Radar plot comparing kinematic features between all novices and experts

**Table 1** Performance of the best Machine Learning models using PCA and without PCA

| | Best model | Best scaler | Accuracy | F1 score | PPV | NPV |
|---|---|---|---|---|---|---|
| **w/o PCA** | SVC | Robust Scaler | 0.774 ± 0.125 | 0.775 ± 0.117 | 0.804 ± 0.110 | 0.630 ± 0.242 |
| **w PCA** | Random Forest | Robust Scaler | 0.795 ± 0.065 | 0.778 ± 0.071 | 0.798 ± 0.076 | 0.698 ± 0.220 |

### 3.2 DAE with Classifiers

Table 2 shows the results from DAE with various classifiers. Though 1-D CNN (0.806) outperformed Random Forest (0.797) on F1 score, Random Forest outweighed its counterpart on accuracy, positive predictive value, and negative predictive value metrics. For a complete set of results for all models, see Table S2 in the Supplementary materials.

**Table 2** Comparison of the best Machine Learning models and 1-D CNN using DAE.

| Best model | Best scaler | Accuracy | F1 score | PPV | NPV |
|---|---|---|---|---|---|
| **Random Forest** | Power Transformer | 0.822 ± 0.072 | 0.797 ± 0.090 | 0.845 ± 0.065 | 0.887 ± 0.174 |
| **1-D CNN** | Robust Scaler | 0.817 ± 0.108 | 0.806 ± 0.110 | 0.829 ± 0.103 | 0.789 ± 0.231 |

## 4 DISCUSSIONS

Our study demonstrates a novel, fully automated pipeline for surgical skill assessment utilizing video-based tool motion data in a laparoscopic fundoplication task. By integrating SAM with YOLOv8 model, we automated the tool segmentation process, eliminating the need for manual annotation, which was a key drawback in prior works[36, 37]. Furthermore, we evaluated both supervised and unsupervised learning approaches, achieving robust performance through classical machine learning on engineered kinematic features including root mean square (RMS) Velocity, RMS Acceleration, RMS Jerk, Path Length, and Bimanual Dexterity as well as leveraging automatically learned features using a DAE.

Recent advances in artificial intelligence have led to the development of a variety of models for automated surgical skill assessment, each with distinct input requirements and levels of interpretability. One approach uses a fully convolutional network (FCN) trained on kinematic data to classify skill levels based on motion patterns, avoiding the need for handcrafted features and offering high interpretability [38]. In contrast, other models operate directly on video data. A 3-D convolutional neural network (3-D CNN) was developed to extract spatiotemporal features from laparoscopic video frames, allowing for autonomous skill classification from visual input alone [39]. While this method captures more complex visual context than FCNs, it requires pre-segmented videos and more computational resources.

Compared to previous work, such as the motion analysis of the JIGSAWS dataset by Lefor et al. [40], our approach extends beyond statistical group comparisons by leveraging machine learning models for classification. While their study provides insights into how kinematic metrics such as path length, velocity, and jerk correlate with expert-annotation skill levels, it is limited to descriptive analytics and manual segmentation of robot-acquired kinematics. In contrast, we provide a scalable strategy that focuses on predictive modeling rather than purely statistical evaluation.

Similarly, the vision-based motion tracking system proposed by the authors of [41] validates the correlation between tool motion features and global rating scale (GRS) scores across standard tasks like suturing and knot tying. However, their setup still involves external tracking systems (hardware-based motion capture tools) and manual data handling, whereas our method leverages existing surgical videos and automates the entire data extraction and labeling pipeline.

When comparing Machine Learning models to DAE for skill assessment, the approach to feature extraction and pattern recognition differs significantly. In traditional Machine Learning models, we explicitly use a variety of features such as distance, displacement, velocity, acceleration, jerk, and bimanual dexterity to identify patterns that reflect skill levels. These features are manually selected and engineered to capture relevant aspects of the task being analyzed.

On the other hand, a DAE operates in an unsupervised manner. It automatically learns a compact, efficient representation of the data without the need for manual feature selection. The autoencoder model discovers and forms these representations based on the inherent structure of the input data. The DAE has been successfully used for summative assessment of participants performing the fundamentals and laparoscopic surgery (FLS) pattern cutting task with high accuracy[42]. In our findings, we showed a substantial improvement in performance with F1 score increasing from 0.778 to 0.806.

An important advantage of our approach is its reliance solely on vision-based tool tracking, without the use of external motion sensors or kinematic data obtained directly from robotic surgical systems. This significantly reduces system complexity and cost, making it more feasible for implementation in real-world surgical training settings. All tool motion data was derived from videos, eliminating the need for specialized equipment and allowing for non-invasive assessment. Though advantageous, our approach currently can track tool motions only in two dimensions as opposed to three-dimensional motion in the real world. In our work, we were able to achieve high performance in correctly classifying the skill level even with two-dimensional motion information. In future, if depth information is needed, it can be readily estimated using deep learning model trained specifically for depth estimation[43, 44].

Our data was originally collected for validating assessment metrics for laparoscopic Nissen Fundoplication performed on a porcine model and hence the data used in this work contains only this technique. Nevertheless, our approach for assessment using kinematic metrics is generalizable to any laparoscopic procedures. Depending on the type of laparoscopic tools, additional fine tuning may need to be performed in our tool tracking model to detect and track new surgical tools. We believe that by extrapolating this system to other procedures we can make a significant impact in the field of surgical skill assessment by eliminating the time and resources constraints currently required for video-based assessment.

## 5 CONCLUSIONS

Our study advances surgical skill assessment by integrating the SAM with a YOLOv8 model trained on the Cholec80 dataset. This integration automates tool tracking and classification, enhancing the precision of surgical training evaluations. Our main contributions are the development of this novel automated system and creating a comprehensive dataset, which improves the accuracy and efficiency of evaluating surgical skills. We compared the traditional approach of using Machine Learning (for supervised learning) with DAE (unsupervised learning) and showed a preference for DAE. This work paves the way for future innovations in surgical education and provides a scalable, efficient assessment tool that aids surgical skill assessment.

## DATA AVAILABLITY

The data that supports the findings of this study may be provided upon reasonable request.

## ACKNOWLEDGEMENTS

We thank Dave Primm of the UT Southwestern Medical Center for help in editing this article.